%% file: bare_jrnl_new_sample4.tex
\documentclass[lettersize,journal]{IEEEtran}
\usepackage{amsmath,amsfonts}
\usepackage{algorithmic}
\usepackage{algorithm}
\usepackage{array}
\usepackage[caption=false,font=normalsize,labelfont=sf,textfont=sf]{subfig}
\usepackage{textcomp}
\usepackage{stfloats}
\usepackage{url}
\usepackage{verbatim}
\usepackage{graphicx}
\usepackage{cite}

\usepackage{amssymb}
\usepackage{booktabs}
\usepackage[dvipsnames,table,xcdraw]{xcolor}
\usepackage{multirow}
\usepackage{colortbl}
\usepackage{color}
\usepackage{tikz}
\usepackage{booktabs}
\usepackage{threeparttable}                                                         
\usepackage{float} 
\usepackage{times}
\usepackage{epsfig}
\usepackage{cuted}
\usepackage{caption}
\usepackage[capitalize]{cleveref}
\begin{document}

\title{Progressive Feature Learning for \\ Realistic Cloth-Changing Gait Recognition}

\author{Xuqian Ren, Saihui Hou, Chunshui Cao, Xu Liu and Yongzhen Huang 
\thanks{Xuqian Ren is with the Computer Science Unit, Faculty of Information Technology and Communication Sciences, Tampere Universities, Tampere 33720, Finland. This work is finished when she is an intern at Watrix Technology Limited Co. Ltd before becoming a Ph.D. candidate at Tampere Universities (E-mail: xuqian.ren@tuni.fi). }
\thanks{Saihui Hou and Yongzhen Huang are with School of Artificial Intelligence,
Beijing Normal University, Beijing 100875, China and also with Watrix Technology Limited Co. Ltd, Beijing 100088, China (E-mail: housaihui@bnu.edu.cn, huangyongzhen@bnu.edu.cn). They are the corresponding authors of this paper.}
\thanks{Chunshui Cao and Xu Liu are with Watrix Technology Limited Co. Ltd, Beijing 100088, China.}}

\markboth{Journal of \LaTeX\ Class Files,~Vol.~14, No.~8, August~2021}%
{Shell \MakeLowercase{\textit{et al.}}: A Sample Article Using IEEEtran.cls for IEEE Journals}

\IEEEpubid{0000--0000/00\$00.00~\copyright~2021 IEEE}

\maketitle

\begin{abstract}
Gait recognition is instrumental in crime prevention and social security, for it can be conducted at a long distance to figure out the identity of persons. 
%
However, existing datasets and methods cannot satisfactorily deal with the most challenging cloth-changing problem in practice. 
%
Specifically, the practical gait models are usually trained on automatically labeled data, in which the sequences' views and cloth conditions of each person have some restrictions.
%
To be concrete, the cross-view sub-dataset only has normal walking condition without cloth-changing, while the cross-cloth sub-dataset has cloth-changing sequences \emph{but only in front views}.
As a result, the cloth-changing accuracy cannot meet practical requirements.
In this work, we formulate the problem as Realistic Cloth-Changing Gait Recognition (abbreviated as RCC-GR) and we construct two benchmarks: CASIA-BN-RCC and OUMVLP-RCC, to simulate the above setting.
Furthermore, we propose a new framework called Progressive Feature Learning that can be applied with off-the-shelf backbones to improve their performance in RCC-GR. 
Specifically, in our framework, we design Progressive Mapping and Progressive Uncertainty to extract cross-view features and then extract cross-cloth features on the basis. 
In this way, the feature from the cross-view sub-dataset can first dominate the feature space and relieve the uneven distribution caused by the adverse effect from the cross-cloth sub-dataset. 
The experiments on our benchmarks show that our framework can effectively improve recognition performance, especially in the cloth-changing conditions. 
\end{abstract}


\section{Introduction}
\label{sec:intro}
\begin{figure}[t]
	\centering	 
	\includegraphics[width=\linewidth]{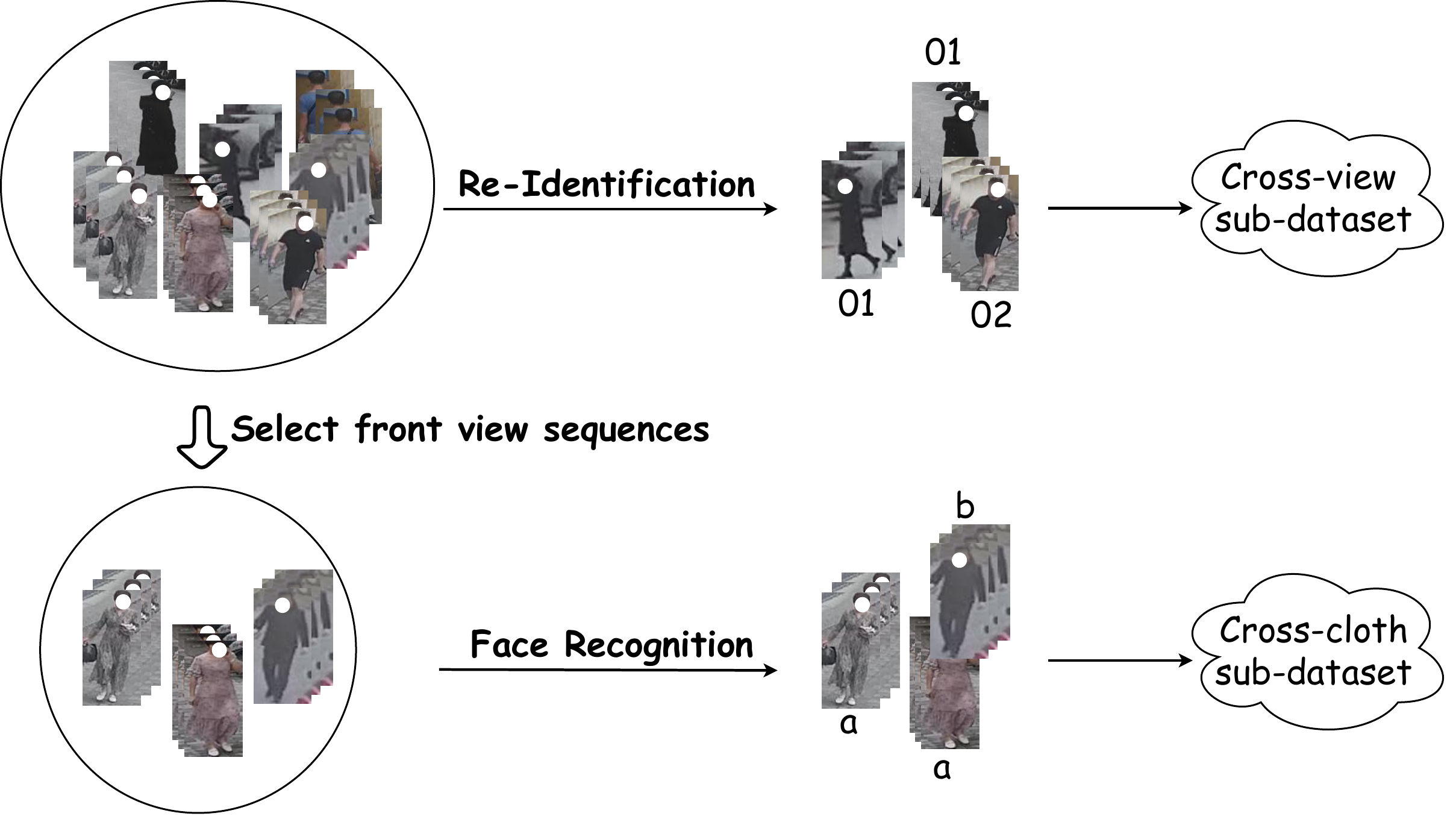}	 	
	\caption{The automatic label pipeline. The images are come from DeepChange dataset~\cite{xu2021long}.}
	\label{fig:label}
        \vspace{-5mm}
\end{figure}

Gait recognition is a valuable tool in video surveillance systems for its outstanding performance in conducting identity verification and abnormal behavior analysis~\cite{larsen2008gait,bouchrika2011using,black2017forensic}.
%
%
However, there still exist a lot of challenges in gait recognition~\cite{yu2006framework,liao2017pose,takemura2018multi,connor2018biometric,9359526}.
Among them, a key challenge is to deal with the cloth-changing condition in real-world applications.

Previous research for gait recognition is mostly conducted with the assumption that the sequences of all views and clothes for each person are available, which, however, \textbf{is nearly impossible in practice} when collecting large labeled datasets, especially in the wild~\cite{zhu2021gait}.
When collecting gait sequences without labeling, it is true that each person's sequences will have multiple views and cloth conditions.
But it is challenging to manually associate sequences of one person.
This requires a significant amount of human effort, which is time-consuming and costly.
Therefore, it is more practical to use automatic methods to assign labels to these sequences in order to get a labeled dataset.
Here we introduce a practical data label method that is usually used to process walking sequences collected in the wild.
\IEEEpubidadjcol

\paragraph{Automatic Label Pipeline}
As illustrated in Figure~\ref{fig:label}, when collecting the gait benchmarks in the wild automatically, it is relatively easier to obtain the cross-view sequences for each person aided by the techniques such as person re-identification (ReID)~\cite{chen2017person,zheng2016person}.
However, currently, ReID mostly relies on features such as the cloth's color and type to relate the video clips of the same person, so this sub-dataset predominantly contains short-term tracks of one person that only has view variations with the same cloth.
In contrast, clustering the cross-cloth sequences of the same person relies on the facial features~\cite{deng2019arcface,deng2020retinaface}, which are only available in the front view.
This is because the front view is the perfect angle to help the cameras capture clear face pictures, so to guarantee label accuracy, only image sequences in the front view will be selected to construct face recognition to get labels.
And since only front view sequences can be used for the cross-cloth sub-dataset, the restrictive requirement leads to the number of sequences we can get being much smaller than those collected for the cross-view sub-dataset.
And in most cases, the identities in the cross-view and cross-cloth subsets have no overlap because we use two different data collection methods to get the two sub-datasets, and the data volume is large.
Although there may be some persons have been collected in both the two sub-datasets, they contribute little to the model to learn cross-view and cross-cloth ability simultaneously because these sequences have different labels in these two sub-datasets.

\begin{figure*}[t]
	\centering	 
	\includegraphics[width=\linewidth]{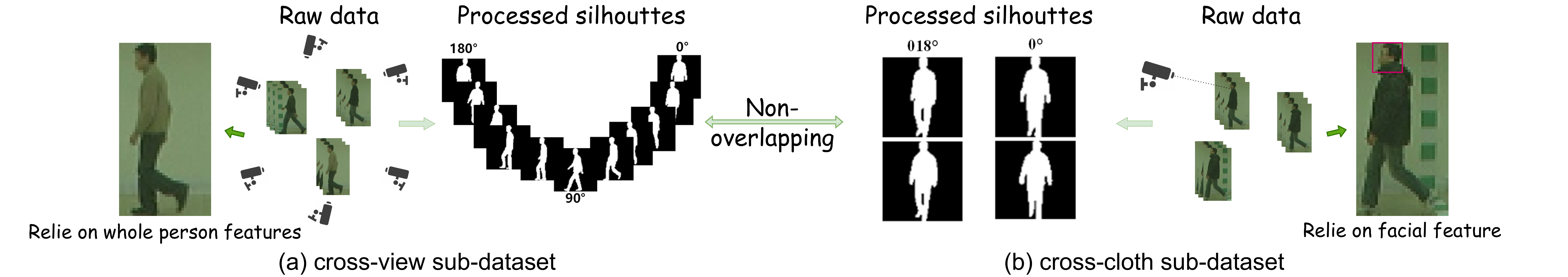}	 	
	\caption{The composition of a real cloth-changing benchmark. (a) Cross-view sub-dataset. Each person has view variations but only has the normal walking condition (NM). (b) Cross-cloth sub-dataset. Each identity has walking in different coats condition (CL) but only has limited views (only front views).
 }
	\label{fig:dataset}
 \vspace{-5mm}
\end{figure*}

\paragraph{Automatically Labeled Data:}
Through this labeling method, the datasets we get have four characteristics:
(1) The sub-dataset that has view variations does not have cloth changing.
(2) The sub-dataset that has cloth variations only has front view sequences.
(3) The volume of the cross-view sub-dataset is larger than the cross-cloth sub-dataset
(4) The two sub-datasets nearly have no overlap.
%
%
%
We show the composition of a realistic cloth-changing dataset in Figure~\ref{fig:dataset}.
It closely resembles the real-world cloth-changing challenges, unlike the current datasets, which are collected from the lab or in the wild but have assigned the labels manually.
In this way, when using automatic data collection methods, it is necessary to develop a new framework to help previously published gait recognition methods perform better on such data. 


In this work, we propose a new task named Realistic Cloth-Changing Gait Recognition (abbreviated as RCC-GR) to simulate the above setting, which is much more meaningful but has not attracted enough attention.
In the following, we will elaborate the benchmark construction\footnote{The benchmark construction is not declared as one of the main contributions, which is only used to facilitate the research.} and an effective approach to deal with the problem respectively.

\paragraph{Benchmark Construction} To support our research, we first construct two benchmarks: \textbf{CASIA-BN-RCC} and \textbf{OUMVLP-RCC}, based on two famous datasets: CASIA-B~\cite{yu2006framework} and OU-MVLP~\cite{takemura2018multi}.
Each benchmark contains two sub-datasets.
The first sub-dataset is used to extract cross-view features, including persons that have view variations, however, they only have the normal walking condition (NM).
And the second sub-dataset is used to extract cross-cloth features, containing other different persons that have the normal walking condition (NM) and walking in different coats condition (CL), while the sequences view are limited in around $0^{\circ}$ and $180^{\circ}$.
Particularly, when we use appearance-based methods, we take silhouettes as the input for gait recognition in this work, so the sequences around $180^{\circ}$ can also be included for the cross-cloth subset due to the horizontal symmetry to enlarge the dataset volume. 
In this way, we obtain two benchmarks that can force the algorithm to realize cross-view and cross-cloth with two sub-datasets, which can simulate real-world applications as much as possible.

\paragraph{Effective Approach}
After observing the statistics of the datasets, we propose a new framework that utilizes progressive feature learning to realize gait recognition.
The framework mainly consists of two methods to map features extracted from the backbone into discriminative feature spaces in a cascaded way, called Progressive Mapping and Progressive Uncertainty, respectively.

Specifically, the previous method used a one-stage module to map features in the cross-view and cross-cloth feature space simultaneously. 
However, in our benchmarks, the cross-view sub-dataset has richer intra-class diversity and has a wider feature span in the high dimensional space, whereas the cross-cloth sub-dataset has a relatively smaller spatial span in the feature space due to its scarcity in intra-class variety.
So if the feature extracted from the two sub-datasets is mapped together, the spatial space will be distorted~\cite{liu2020deep}.
Therefore, our Progressive Mapping applies a two-stage module, which contains a Cross-View Mapping module (CVM) and a Cross-Cloth Mapping module (CCM) to fully span the cross-view feature in the spatial space first, and then extract the cross-cloth feature on the basis.
In this fashion, we relieve the uneven distribution in the feature space.

Besides, the Progressive Uncertainty maps each sequence as a Cross-View Gaussian distribution and a Cross-Cloth Gaussian distribution instead of a point.
Particularly, mapping each sequence as Gaussian distributions allows the network to increase gradient weight from easy learning cross-view sequences and prevent the network from overfitting on the hard cross-cloth sequences.
Furthermore, by re-sampling new embeddings from the Gaussian distributions, we can gain more knowledge from the sequence that can be used to classify each identity. 
Therefore, mapping sequences as Gaussian distributions can further make the feature extracted from the cross-view sub-dataset have more effects in model training.

\paragraph{Contributions} To sum up, our contributions mainly lie in three folds:
\begin{itemize}
	\item[$\bullet$]  We propose a new setting to simulate the cloth-changing gait recognition problem in practice, which is needed to realize cross-view and cross-cloth with two non-overlapping sub-datasets.
	\item[$\bullet$]  We propose a new framework that utilizes progressive feature learning to fully expand the gait representations in discriminative space. In this framework, we design a Progressive Mapping and a Progressive Uncertainty to extract features used for cross-view and cross-cloth in a cascaded way. 
	\item[$\bullet$] Extensive experiments on our benchmarks prove the effectiveness and outstanding performance of our proposed framework for the cloth-changing gait recognition problem in practice. 	
\end{itemize}

Our framework can be conveniently employed with off-the-shelf backbones, and we believe our work is crucial to the research of gait recognition in real-world applications as well as serving as a foundation stone for further advanced methods.

\section{Related Work}
\label{sec_related_work}
\subsection{Gait Recognition}
\noindent\textbf{Model-based Gait Recognition:} 
These methods~\cite{liao2020model,li2020end,li2021end,li2022strong,9478261} aim to extract human pose parameters that are more invariant to the view, clothing, and carrying variations from videos to realize gait recognition. 
PoseGait~\cite{liao2020model} exploits the 3D coordinates of joints of the human body as features and designs some handcrafted features based on the 3D pose. Then, CNN extracts the spatio-temporal information from the handcrafted features. 
CycleGait~\cite{li2022strong} utilizes gait periodicity priors to extract temporal features, thereby reducing the need for precise gait cycle segmentation.
%
%
%
These methods give gait recognition physical explanation and are more suitable with high-resolution inputs.

\noindent \textbf{Appearance-based Gait Recognition:} 
These methods aim to extract discriminative and robust representations from silhouettes. 
They can perform recognition at lower resolutions and are suitable for outdoor applications. 
According to the input of the network, the appearance-based model can be roughly further divided into three categories: template-based method~\cite{shiraga2016geinet,hu2018robust,he2018multi}, video-based method~\cite{liu2018learning,wolf2016multi,8643814,lin2020gait,lin2021gait,9767609} or image set-based method~\cite{chao2019gaitset,fan2020gaitpart,hou2020gait,hou2021set}. 
Here we show some representative work of each category. 
1) Based on templates: 
GEINet~\cite{shiraga2016geinet} uses gait energy image (GEI), the most prevalent image-based gait representation, as input and uses a CNN to extract its potential gait patterns. 
%
%
%
However, the template cannot encode the temporal information completely, leading to sub-optimal performance.
2) Based on videos: 
Video-based methods usually use 3D convolution to learn sequence features from both the spatial and temporal dimensions.
%
%
Wolf \textit{et al.}~\cite{wolf2016multi} uses 3D CNN to extract spatio-temporal information from videos with a fixed-length. 
To further deal with the variant length of sequences, MT3D~\cite{lin2020gait} uses a frame pooling algorithm to make the input sequences have the same length and designs two branches to extract the spatial-temporal information at multiple scales. 
Also, GaitGL~\cite{lin2021gait} uses 3D CNN to attain discriminative representations from both global and local features by a Global and Local Feature Extractor.
3) Based on image sets: 
GaitSet~\cite{chao2019gaitset} argues that the order of the gait silhouettes is not vital to classify each person and first proposes to extract frame-level and set-level features from an unordered set without ranking each frame sequentially like in a video. 
%
%
GaitPart~\cite{fan2020gaitpart} further enhances the part-level features with a Focal Convolution Layer and a Micro-motion Capture Module. 
%
%
GLN~\cite{hou2020gait} extracts frame-level and set-level features from different stages and merges them by a lateral connection. It also uses a Compact Block to compress gait representations. 
%

\subsection{Data Uncertainty Learning}
To improve the robustness and interpretability of CNN, data uncertainty learning has been gradually used in face recognition~\cite{shi2019probabilistic} and person re-identification~\cite{yu2019robust,9707634,9483643} and other computer vision fields~\cite{8693882,9439889,9882310,9899753,9966825} to help the network reject noisy input and avoid false recognition. 
The feature learned by CNN directly may be ambiguous and may not even be present in the input. 
So to model this uncertainty of learned feature, this kind of method chooses to model each sequence as a Gaussian distribution instead of a single point.
With the help of confidence, the network can reject noisy input and avoid false recognition.

\noindent\textbf{Data uncertainty learning in face recognition:} 
PFE~\cite{shi2019probabilistic} first proposes to map each face image as a Gaussian distribution, regarding the sequence feature as the mean, and adding another branch to learn the confidence for the sequence feature. 
The mean of the distribution can be regarded as the most likely feature of the sequence mapped in the latent space, and the variance can be seen as the span of the distribution that in what range the feature can be seen as a mapping from the sequence. 
DUL~\cite{chang2020data} expands PFE to learn the feature and uncertainty simultaneously. Therefore, the learned uncertainty can affect feature learning by adaptively reducing the negative influence of noisy training samples. 
Shi \textit{et al.}\cite{shi2020towards} learns a universal representation by splitting the feature representation into sub-embeddings and relaxing the confidence to be sub-embedding specific. 
ProbFace~\cite{chen2021reliable} further re-designs the triplet loss, making it aware of the uncertainty, which can be used in our gait recognition task. The triplet loss can help our network focus more on the data from cross-view sub-dataset, which is the normal walking condition in our task.

\noindent\textbf{Data uncertainty learning in Re-ID:} 
DistributionNet~\cite{yu2019robust} uses uncertainty learning to deal with noisy samples, such as the samples with the wrong label or outliers' data. 
SCFU~\cite{shi2021spatial} exploits spatial-wise and channel-wise uncertainty to solve the occluded person re-identification problem. 
Inspired by their works, we use uncertainty learning to solve the cross-cloth gait recognition problem at both the distribution and feature levels. 
The variance generated by our framework can model the optimization difficulty caused by the quantity imbalance and silhouettes variation diversity. 
In this way, the class lacking intra-class diversity and the parts with more variations in each feature will have larger variance and contribute less to the gradient, which makes the model learn a robust template from the classes that have rich feature diversity.

\section{Our Approach}
\label{method}
In this paper, we propose a new task called \textbf{Realistic Cloth-Changing Gait Recognition}, which focuses on the cloth-changing problem in practice.
In this section, we will formally define the problem and then introduce a framework with Progressive Feature Learning to fully extract the cross-view and cross-cloth information in realistic datasets.
Specifically, we will describe the details of the two new methods we used in our framework: Progressive Mapping and Progressive Uncertainty.


\subsection{Problem Definition}
We denote the whole training set as $(\mathcal{X}_v \cup \mathcal{X}_c)$ to represent the two sub-datasets used for cross-view and cross-cloth separately.
In $\mathcal{X}_v$, each person only has NM condition, while in $\mathcal{X}_c$, NM and CL sequences around front views can be used.
It should be noted that $\mathcal{X}_v$ and $\mathcal{X}_c$ do not have any overlap.
And in most cases, the number of sequences in $\mathcal{X}_v$ is much larger than that in $\mathcal{X}_c$, since it is much easier to collect cross-view data aided by ReID.
In the test phase, sequences in all conditions and views will be used for evaluation.
So when considering the properties of the proposed benchmarks, we need to alleviate the imbalance in datasets and fully explore the cross-view and cross-cloth information.

\subsection{Framework Overview }

To solve this realistic cloth-changing task, we propose a framework that utilizes Progressive Feature Learning.
Our framework can be applied to both appearance-based and model-based methods.
To simplify the illustration, we take silhouettes as input which is more suitable for realistic low-resolution conditions.

\begin{figure*}[t]
	\centering	 
	\includegraphics[width=\linewidth]{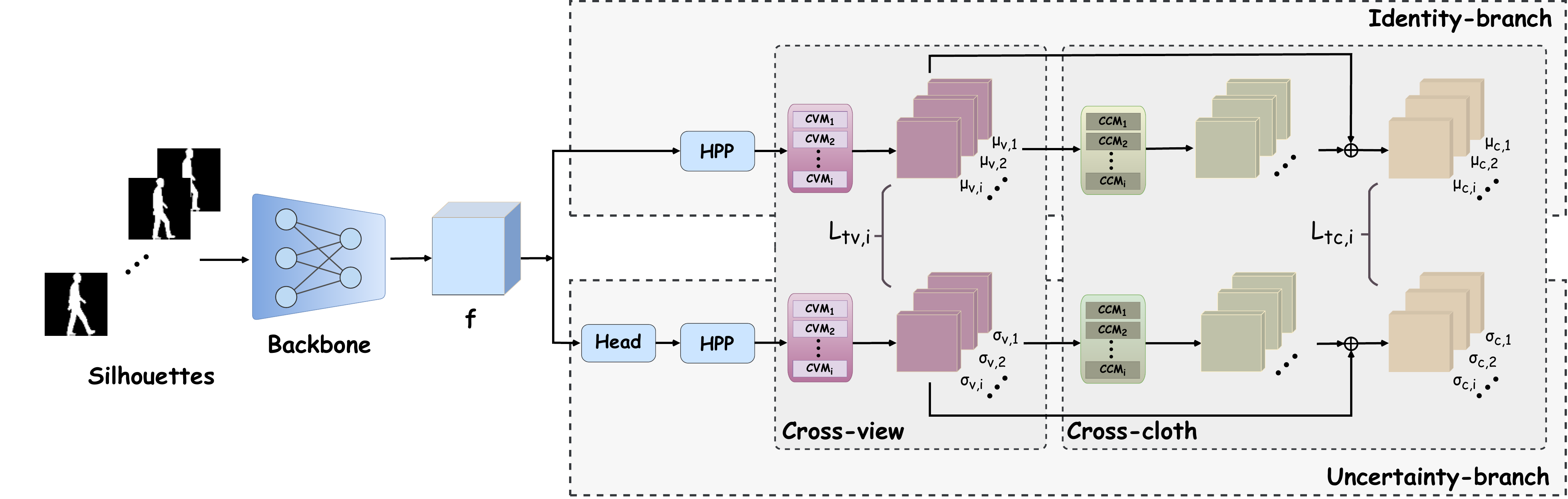}	 	
	\caption{The structure of our framework. Backbone is used to extract features from silhouettes. The feature $f$ outputs from the backbone will be put into two branches, and each uses Progressive Mapping with Cross-View Mapping module (CVM) and Cross-Cloth Mapping module (CCM) to generate $\mu_{j,i}/\sigma_{j,i};\ j \in v,c; \ i \in 1,2,...,S$. \textit{Head} is a light block used to isolate features of two branches. Progressive Uncertainty constructs two Gaussian distributions with learned mean and variance features to relax constraints on feature mapping. }
	\label{fig:structure}
 \vspace{-5mm}
\end{figure*}

According to the sequence quantity proportion in $\mathcal{X}_v$ and $\mathcal{X}_c$, there is a data imbalance in benchmarks, which will make the model biased if it learn the cross-view and cross-cloth ability at the same time.
If the cross-view feature and cross-cloth feature are treated equally, the uneven data distribution will make them hard to separate from each other and distort feature spaces.
So, making features from $\mathcal{X}_v$ domain the feature space first can train the model more robust.
To achieve this, we design Progressive Mapping and Progressive Uncertainty in our framework, and the overview is shown in Figure~\ref{fig:structure}.
The training pipeline is as followings.
First, the feature extracted from a sequence through the backbone is named $f$.
$f$ will be transferred into two branches, called the identity-branch and uncertainty-branch separately.
In each branch, we employ Progressive Mapping method, which uses a Cross-View Mapping module (CVM) and a Cross-Cloth Mapping module (CCM), to first extract the cross-view features from $\mathcal{X}_v$, called $\mu_v/\sigma_v$, and then also extract the cross-cloth features $\mu_c/\sigma_c$ from $\mathcal{X}_c$ on the basis.
Next, Progressive Uncertainty will map each sequence as a cross-view Gaussian distribution $\mathcal{N}\left(\mu_{v}, \sigma_{v}^{2} \mathrm{I}\right)$ and a cross-cloth Gaussian distribution $\mathcal{N}\left(\mu_{c}, \sigma_{c}^{2} \mathrm{I}\right)$ in feature spaces.
Mapping features as Gaussian distributions can make the feature extracted from the $\mathcal{X}_v$ domain the feature space first and then span cross-cloth features from $\mathcal{X}_c$.
Finally, Uncertainty-aware Triplet Loss~\cite{chen2021reliable} $\mathcal{L} _{tv}$ and $\mathcal{L} _{tc}$ are employed to train the whole framework.

\subsection{Progressive Mapping}
\label{sec:progmapping}

Progressive Mapping aims to make features from $\mathcal{X}_v$ domain the feature space first, making the model learn to separate each identity better.
This method has two key components: A two-stage module and Progressive-aware Triplets.

\noindent\textbf{Two-stage module:}
A two-stage module is designed to span the feature used for cross-view and cross-cloth in a cascaded way.
Like in GaitSet~\cite{chao2019gaitset} and GaitPart~\cite{fan2020gaitpart}, in each branch, before the two-stage module, a Horizontal Pyramid Pooling (HPP)~\cite{chao2019gaitset} is employed to first slice the feature horizontally into $S$ parts:
\begin{equation}
    z_i = HPP(f_i); \ i \in 1,2,...,S
\end{equation}
Then, the two-stage module accepts each part of the feature and inputs them to Cross-View Mapping module (CVM) with $S$ parts, which is used to first span the feature extracted from $\mathcal{X}_v$, and then another Cross-Cloth Mapping module (CCM) with $S$ parts is concatenated to span the feature extracted from $\mathcal{X}_c$ on the basis. 
Each module of CVM and CCM works in parallel for each part of the feature, $z_i$.
Also, a residual connection between the output of CVM and CCM is added to avoid the loss of cross-view information.
CCM extracts the cross-cloth feature on the basis of the cross-view feature, which alleviates the pressure on models that are trained to extract the cross-view and the cross-cloth feature simultaneously.
We take the identity-branch as an example and define the module with the following formulation:
\begin{equation}
	\mu _{v,i} = CVM_{i}(z_i)
\end{equation}
\begin{equation}
	\mu _{c,i} = \mu _{v,i} + CCM_i( \mu _{v,i} )
\end{equation}
where $\mu_{v,i}$ is the output from CVM, and $\mu_{c,i}$ is the addition result of the output from CVM and CCM. 
In this way, the two-stage module can help the model first to learn the identity classification ability better and then improve its cross-cloth ability.


\noindent\textbf{Progressive-aware Triplets:} Based on the two-stage module, we need to re-design the triplets' selection method to make the performance of the two-stage module maximum. 
We denote a sample triplet as $(a,p,n)$, where $a$ denotes the anchor, $p$ denotes a sample which label is as same as the anchor $a$, and $n$ is a sample which labels are different from that of $a$. 
In our work, we only choose persons from $\mathcal{X}_v$ in each batch to construct $\mathcal{T}_v = (a,p,n)$ when calculating the loss behind CVM, making it only extract the cross-view feature. 
Then we use the whole batch to construct $\mathcal{T}_c = (a,p,n)$ when calculating loss behind CCM, so the CCM can make full use of the whole information to cross-cloth as well as maintain the cross-view ability. 
Also, like in previous research, we select $p \times k$ sequences in each iteration, where $p$ represents the number of persons and $k$ is the number of training sequences each person has in the batch.
To make the triplets maximum in cross-cloth triplets as well as guarantee the equality, we choose to randomly select $(p/2,k)$ batch from $\mathcal{X}_v$, and $(p/2,k)$ batch from $\mathcal{X}_c$ with $k/2$ from the NM and $k/2$ from CL. 
More specifically, for CASIA-BN-RCC, the $(a,p,n)$ are chosen from the 1-44id for $\mathcal{L} _{tv}$, and the $(a,p,n)$ are chosen from the 1-74id for $\mathcal{L} _{tc}$; for OUMVLP-RCC, the $(a,p,n)$ are chosen from the first 3200ids of the training dataset for $\mathcal{L} _{tv}$, and the $(a,p,n)$ are chosen from the whole 5143id from training set for $\mathcal{L} _{tc}$.
The details about the loss can be seen in Section~\ref{sec:proguncertainty}.
In this way, we force the network to first learn features from $\mathcal{X}_v$ and then extract features from both  $\mathcal{X}_v$ and $\mathcal{X}_c$.

\subsection{Progressive Uncertainty}
\label{sec:proguncertainty}
Progressive Uncertainty maps each sequence as a Cross-View Gaussian distribution and a Cross-Cloth Gaussian distribution in the feature spaces instead of a point, which can relax the constraints for the model.
The identity-branch accepts features output from the backbone and learns the mean feature, which is used to represent the characteristic of the gait sequence and will be used during inference. 
%
%
The uncertainty-branch is used to generate the variance feature, representing with what uncertainty a feature can represent this sequence, and this branch will be abandoned during inference.
In this branch, the original feature $f$ outputs from the backbone will first pass through a Head module, which is a light block with several convolution layers, to make the feature isolated from the identity branch.
Then, like the identity-branch, a two-stage module is employed in the Uncertainty-branch to generate two variance features as followings:
\begin{equation}
	\sigma _{v,i}=Relu\left( CVM_i(HPP\left( Head( f  \right)) )\right) 
\end{equation}
\begin{equation}
	\sigma _{c,i}=Relu\left(\sigma _{v,i}+CCM_i\left( \sigma _{v,i} \right) \right) 
\end{equation}

where $\sigma _{v,i}$ is the output from each module in CVM and $\sigma _{c,i}$ is the result that added the outputs from CVM and CCM. 
To ensure the variances are positive, we apply the Rectified linear unit.
With mean feature $\mu_{v,i}$ and $\mu_{c,i}$ and variance feature $\sigma_{v,i}$ and $\sigma_{c,i}$, we construct Cross-View Gaussian distributions $\mathcal{N}\left(\mu_{v,i}, \sigma_{v,i}^{2} \mathrm{I}\right)$ and Cross-Cloth Gaussian distributions $\mathcal{N}\left(\mu_{c,i}, \sigma_{c,i}^{2} \mathrm{I}\right)$.
In this way, the representation of each sequence is not point embeddings but stochastic embeddings sampled from $\mathcal{N}\left(\mu_{j,i}, \sigma_{j,i}^{2} \mathrm{I}\right), j\in v,c$.
$\mu_{j,i}$ used during inference can be regarded as sampled from the center point of the Gaussian distributions.
The variances measure the uncertainty of the classification result of each stage for the Progressive Mapping, which indirectly benefits the supervision information of the label and optimizes the training process~\cite{shi2021spatial}.

\noindent\textbf{The re-parameterization trick:} 
To enrich the information that can be used in classification, we can also re-sample a new embedding from each Gaussian distribution. 
However, the sampling operation mentioned above is not differentiable and prevents the backpropagation of the gradients flow during training~\cite{chang2020data}. 
To solve the problem, we employ a re-parameterization trick~\cite{kingma2013auto} to let the model backpropagate normally like in~\cite{yu2019robust,chang2020data,shi2021spatial}. 
A random noise $\epsilon$, which is independent of the model parameters, is sampled from a normal distribution and then used to generate the embedding $e_{j,i}$, making them to be regarded as re-sampled representations from Gaussian distributions mapped by a sequence.
\begin{equation}
	e_{j,i}=\mu_{j,i} +\varepsilon \sigma_{j,i} ;\  \epsilon \sim \mathcal{N}(0, \mathrm{I}); \ j \in v,c;\ s \in 1,2,...,S
\end{equation}
where $e_{j,i}$ is the embedding re-sampled with the help of $\mu_{j,i}$ and $\sigma_{j,i}$.

\noindent\textbf{Uncertainty-aware Triplet Loss:} 
The mean feature $\mu_{j,i}$, variance feature $\sigma_{j,i}$, and the sampled embedding $e_{j,i}$ are used to control the triplet loss. 
We employ an uncertainty-aware triplet loss~\cite{chen2021reliable} after CVM and CCM separately:

\begin{equation}
    \begin{aligned}
  \mathcal{L} _{tv,i}=&\frac{1}{|\mathcal{T}_v |}\sum_{(a,p,n)\in \mathcal{T}_v}{\left[ \frac{\left\| \mu _{v,i}^a-\mu _{v,i}^p \right\| ^2}{\sigma _{v,i}^{a^2}+\sigma _{v,i}^{p^2} }-\frac{\left\| \mu _{v,i}^a-\mu _{v,i}^n \right\| ^2}{\sigma _{v,i}^{a^2}+\sigma _{v,i}^{n^2}}+m \right] _+}
    \\
    \mathcal{L} _{tc,i}=&\frac{1}{|\mathcal{T}_c |}\sum_{(a,p,n)\in \mathcal{T}_c}{\left[ \frac{\left\| \mu _{c,i}^a-\mu _{c,i}^p \right\| ^2}{\sigma _{c,i}^{a^2}+\sigma _{c,i}^{p^2}}-\frac{\left\| \mu _{c,i}^a-\mu _{c,i}^n \right\| ^2}{\sigma _{c,i}^{a^2}+\sigma _{c,i}^{n^2}}+m \right] _+}   
    \end{aligned}
\end{equation}	

%

where $|\mathcal{T}_i |,i=v,c$ is the number of triplets. 
We can also calculate $\mathcal{L} _{tv,i}^{e}$ and $\mathcal{L} _{tv,i}^{e}$ by replacing $\mu_{v,i}$ and $\mu_{c,i}$ with $e_{v,i}$ and $e_{c,i}$. 

In total, the final loss in our framework can be represent as:
\begin{equation}
	\mathcal{L} =  \mathbb{E}(\sum_{i=0}^S \mathcal{L}_{tv,i}+\mathcal{L} _{tc,i}+\mathcal{L} _{tv,i}^{e}+\mathcal{L} _{tc,i}^{e})
\end{equation}
$\mathbb{E}(\cdot)$ means we sum each triplet loss and average them to gain the final loss.

\section{Experiment}
In this section, we employ two representative backbones: GaitSet~\cite{chao2019gaitset} and GaitGL~\cite{lin2021gait} to demonstrate the effectiveness of our framework. The backbone based on image sets input and the later is based on video input.
It should be noted that our framework can be adopted with both model-based or appearance-based backbones to improve their accuracy in RCC-GR task.
All models are implemented with PyTorch~\cite{paszke2019pytorch}, and we train with CASIA-BN-RCC on TITAN-XP and trained with OUMVLP-RCC on V100 GPUs.

\subsection{Datasets Construction}
\label{details}

\subsubsection{Datasets}
To support our proposed task, we construct two benchmarks CASIA-BN-RCC and OUMVLP-RCC based on two popular gait datasets: CASIA-B~\cite{yu2006framework} and OU-MVLP~\cite{takemura2018multi}.
To construct our benchmarks, we follow 3 principles:
1) The training set collected in real scenes has a sub-dataset that only has NM conditions with view variations, and a sub-dataset has NM/CL conditions only in front views.
2) The two cross-view and cross-cloth sub-datasets have no overlap.
3) The sequence number in the cross-view sub-dataset should be larger than that in the cross-cloth sub-dataset.
%
%

\noindent\textbf{CASIA-BN-RCC} 
The original CASIA-B is a typical gait dataset that consists of 124 identities. 
Each person contains three walking conditions: normal walking (NM-1,2,3,4,5,6), carrying bags (BG-1,2), and walking with different coats (CL-1,2). 
Each walking condition is uniformly distributed in 11 views, from $0^{\circ}$ to $180^{\circ}$. 
%
%
Based on previous research~\cite{chao2019gaitset,fan2020gaitpart}, the first 74 identities are used as the training dataset, and the remaining 50 identities are used for evaluation.
And during the test, NM-1,2,3,4 are grouped into the gallery, and NM-5,6; BG-1,2; CL-1,2 are divided into three subsets according to the waking conditions and are regarded as the probe. 
However, since the segmentation of CASIA-B is relatively rough, we collected some pedestrian images and trained a new segmentation model to re-segment CASIA-B, and gained CASIA-BN. 

In order to construct two sub-datasets for CASIA-BN, we first extract the NM, BG\footnote{Since only the middle part of silhouettes of BG is different with NM in appearance, we regard BG as NM and also extract BG sequences to enlarge the dataset} sequences with all views in 1-44 identities to construct $\mathcal{X}_v$, then extract sequences in 45-74 identities, which walking conditions are NM, BG, CL but views are limited in $0^{\circ}/18^{\circ}/162^{\circ}/180^{\circ}$ to construct $\mathcal{X}_c$.  
The sequence number ratio in the two sub-datasets is 3:1.
The test persons have remained the same as in the previous setting~\cite{chao2019gaitset,fan2020gaitpart}.

\noindent\textbf{OUMVLP-RCC} 
OU-MVLP is a large dataset in public which consists of 10307 identities (5153 identities for training and the rest 5154 identities for testing).
Each person contains 14 views, ranging from $0^{\circ}$ to $90^{\circ}$, $180^{\circ}$ to $270^{\circ}$. 
However, each person only contains one walking condition: normal walking (NM-00,01). 
During the test, NM-01 is regarded as the gallery, and NM-00 is regarded as probe. 
To construct $\mathcal{X}_v$ in OUMVLP-RCC, we extract all sequences of the first 3200 identities in the training dataset. 
%
%
Then for $\mathcal{X}_c$, we extract the sequences which views fall in $0^{\circ}/15^{\circ}/180^{\circ}/195^{\circ}$ in the last 1954 identities and abandon some identities without these views, finally get 1944 IDs.
To try our best to simulate the condition of walking in different coats, we dilate NM-00 to construct CL-00, dilate NM-01 to construct CL-01.
The dilation method is that we dilate the upper part body of each frame with a $5\times 5$ kernel and random kernel type (rect, cross, ellipse) and we define the edge of the upper part of body: upper bound: [10, 14], lower bound: [38, 42], the upper part is selected randomly from this region.
Generate cloth-changing images with dilation can make the dataset distribution similar with CASIA-BN.
The ratio number between $\mathcal{X}_v$ and $\mathcal{X}_c$ also constraints to 3:1.
For evaluation, NM-00 is dilated to construct CL-00, and NM-01 is regarded as the gallery, NM-00 and CL-00 are used as probe in different walking conditions with 14 views.

The statistics of the two benchmarks are shown in Table~\ref{tab:dataset}.
Since we only extract a part number of sequences from original datasets, \textbf{the volume of the new benchmarks is much smaller than the original ones}.
\begin{table}
\begin{center}
\setlength{\tabcolsep}{0.6mm}
\caption{Our constructed benchmarks statistics. NM for \textit{normal walking}, BG for \textit{walking with bags}, CL for \textit{walking in different clothes}.}
 \resizebox{\columnwidth}{!}{%
	\begin{tabular}{ccccccccc}
		\hline
		\multirow{2}{*}{Dataset}      & \multicolumn{2}{c}{Identities}                                   & \multicolumn{5}{c}{Walking condition} & \multirow{2}{*}{View} \\ \cline{2-8}
		& \multicolumn{1}{c}{Train}               & Test                & ID num    & Seq num          & NM    & BG    & CL   &                       \\ \hline
		\multirow{3}{*}{CASIA-BN-RCC} & \multicolumn{1}{c}{\multirow{3}{*}{74}} & \multirow{3}{*}{50} & 44ids   &  3778      & 6     & 2     & -    & 11                    \\ 
		& \multicolumn{1}{c}{}                    &                     & 30ids   &  1195      & 6     & 2     & 2    & 4                     \\  
		& \multicolumn{1}{c}{}                    &                     & 50ids   &  5474   & 6     & 2     & 2    & 11                    \\ \hline
		\multirow{3}{*}{OUMVLP-RCC} & \multicolumn{1}{c}{\multirow{3}{*}{5143}} & \multirow{3}{*}{5154} & 3200ids & 84694 & 2 & - & - & 14 \\  
		& \multicolumn{1}{c}{}                    &                     & 1944ids  & 27750  & 2     & -     & 2    & 4                     \\  
		& \multicolumn{1}{c}{}                    &                     & 5154ids  & 132108  & 2     & -     & 1    & 14                    \\ \hline
\end{tabular}
}

\label{tab:dataset} 
\end{center}
\vspace{-5mm}
\end{table}

\begin{figure}[h]
	\centering	
	\includegraphics[width=\columnwidth]{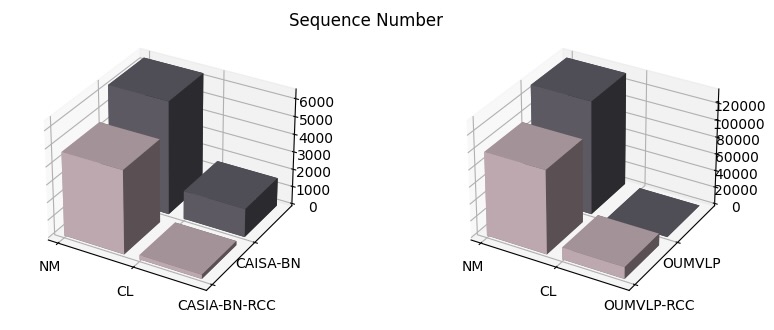}		
        \caption{The comparision of sequence number of each cloth condition between different datasets.}
         \label{fig:volume}
         \vspace{-3mm}
 \end{figure}

The sequence number of each cloth condition in original datasets and our benchmarks used for training can be seen in Figure~\ref{fig:volume}. It can be seen that our benchmarks have a relatively smaller dataset volume compared with previous datasets, but the accuracy can be improved with further collected data.

\begin{figure}[h]
	\centering	
	\includegraphics[width=0.6\columnwidth]{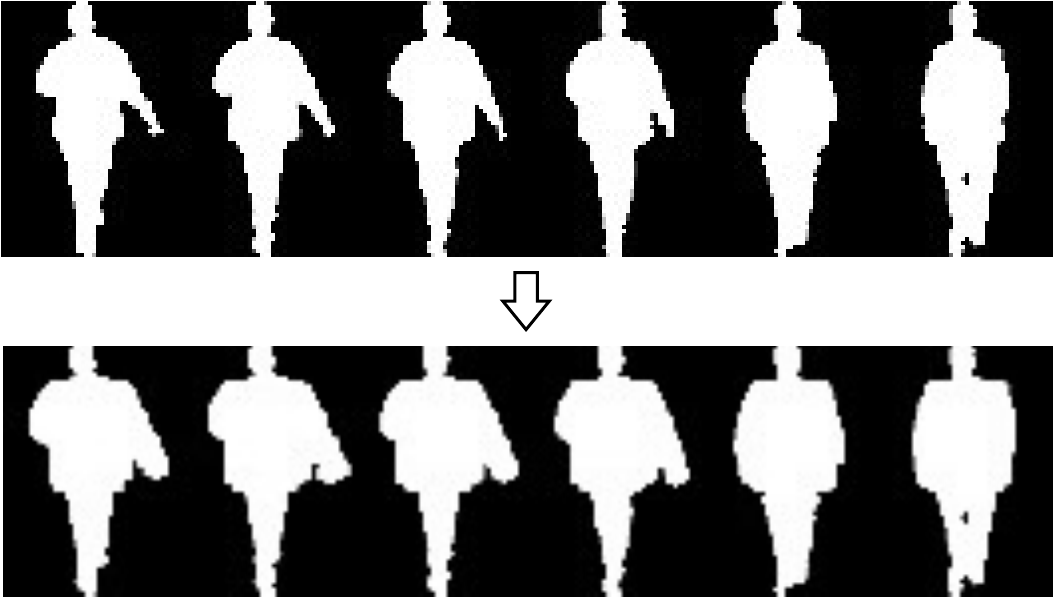}		
    \caption{Some examples of the generated cloth-changing images. The first line shows the original images, the second line shows the generated image after dilation.}
    \label{fig:dilation}
    \vspace{-3mm}
 \end{figure}

 We also show some examples of the generated cloth-changing image in OUMVLP-RCC in Figure~\ref{fig:dilation}. Since we select the upper part of body and dilation methods randomly, each image has higher diversity after dilation.

\input{CASIA-BN2-experiment}

\input{OUMVLP-experiment}

\begin{table}[]
\begin{center}
		\caption{The performance of Progressive Mapping. The results are reported in the rank-1 accuracy. T-S: \textit{two-stage module}. P-T: \textit{Progressive-aware Triplets}. Without P-T, we select triples from both $\mathcal{X}_v$ and $\mathcal{X}_c$.}
        \setlength{\tabcolsep}{0.8mm}
        \begin{tabular}{ccccccccc}
        \hline
        \multicolumn{4}{c}{Settings}                                               & \multicolumn{3}{c}{CASIA-BN-RCC} & \multicolumn{2}{c}{OUMVLP-RCC} \\ \hline
        \multicolumn{1}{c}{T-S} & \multicolumn{1}{c}{$L_{nv,i}$} & \multicolumn{1}{c}{$L_{nc,i}$} & P-T& NM $\uparrow$  & BG $\uparrow$  & CL $\uparrow$  & NM $\uparrow$  & CL $\uparrow$  \\ \hline
                      &  & $\checkmark$ &  & 89.7 & 82.0 & 45.2 & 84.6 & 54.4  \\  
         $\checkmark$ &  & $\checkmark$  &  & 88.8 & 81.6  & 45.9    &   84.0 & 54.4                   \\
         $\checkmark$& $\checkmark$  & $\checkmark$  & & 89.0          & 82.0  & 46.0    & 84.1    & 54.8         \\ 
         $\checkmark$  &  $\checkmark$ & $\checkmark$ & $\checkmark$    & \textbf{89.7}  & \textbf{82.6} & \textbf{48.5} &\textbf{84.7}  &\textbf{ 55.6} \\ \hline
        \end{tabular}%
\label{tab:sep}
\vspace{-5mm}
\end{center}
\end{table}

\subsection{Implementation Details}
\subsubsection{Input}  
The input size of each silhouette is set to $64 \times 64$ for CASIA-BN-RCC and OUMVLP-RCC. 
During training, we select 30 silhouettes randomly from each sequence as input for GaitSet, and select frames in a fixed order for GaitGL. 
The batch size is set as (16,16) for CASIA-BN-RCC and set as (32,8) for OUMVLP-RCC. 
During test, all silhouettes of each sequence are put into model to get their sequence representations. 
If the probe has the lowest Euclidean distance on all parts with one gallery sequence, it will be assigned with the corresponding gallery label.

\subsubsection{Network Structure}
The Network structure we employed for GaitSet~\cite{chao2019gaitset} and GaitGL~\cite{lin2021gait} backbone is shown in Table~\ref{tab:optimization}. 
For the backbone from GaitSet, we employ the simplistic backbone without the Multilayer Global Pipeling~\cite{chao2019gaitset}, which contains three blocks, and each contains two basic convolution layers. 
For the backbone from GaitGL, we use the original backbone proposed in~\cite{lin2021gait}. 
The input and output channels we use for backbones \textbf{are halved and set as (16,32,64)} for GaitSet and GaitGL backbones for CASIA-BN-RCC, for we have less data volume.
The channels for each layer we set for OUMVLP-RCC are (64,128,256).
For the two backbones, the output dimension of the CVM and CCM module used to generate mean features and variance features is set to 64 for CASIA-BN-RCC and 256 for OUMVLP-RCC. 
For simplicity, we employ a Fully Connected Layer (FC) for each CVM and CCM module.
It should be noted that FC is not the only choice, but we employ it for its simplicity and ease of implementation.
And in the head module, we employ two convolution layers (Conv), and each kernel size is $3 \times 3$, padding is 1, and following the first Conv is the Sigmoid function, following the second Conv is the ReLU.
In HPP, we adopt the scale $S={16}$ to split the feature horizontally.

\begin{table*}[h]
\centering

\caption{The optimization method for each backbone used in each dataset. IB means that\textit{ only train with the identity-branch}, UB means \textit{load the pre-trained model and train with the uncertainty-branch}. Total means \textit{the total iteration}. LR means \textit{the learning rate at the beginning}. }
\setlength{\tabcolsep}{1.5mm}
\begin{tabular}{c|c|c|c|c|c|c|c}
\toprule
Backbone & Dataset & Backbone & Branch & MileStone & Total & LR & Weight decay \\ \midrule
\multirow{6}{*}{GaitSet} & \multirow{3}{*}{CASIA-BN-RCC} & \multirow{3}{*}{(16,32,64)} & Upper & (1k,2k,3k) & 35k & \multirow{3}{*}{1e-1} & \multirow{3}{*}{5e-4} \\
 &  &  & IB & (5k,10k,15k) & 20k &  &  \\
 &  &  & UB & (4k,8k,12k) & 15k &  &  \\ \cline{2-8} 
 & \multirow{3}{*}{OUMVLP-RCC} & \multirow{3}{*}{(64,128,256)} & Upper & (50k,100k,125k) & 150k & \multirow{2}{*}{1e-1} & \multirow{3}{*}{1e-4} \\
 &  &  & IB & (50k,100k,125k) & 150k &  &  \\
 &  &  & UB & (50k,80k) & 120k & 1e-2 &  \\ \midrule
\multirow{6}{*}{GaitGL} & \multirow{3}{*}{CASIA-BN-RCC} & \multirow{3}{*}{(16,32,64)} & Upper & 70k & 80k & \multirow{3}{*}{1e-1} & \multirow{3}{*}{5e-4} \\
 &  &  & IB & (10k,20k,30k) & 35k &  &  \\
 &  &  & UB & (3k,5k,8k) & 10k &  &  \\ \cline{2-8} 
 & \multirow{3}{*}{OUMVLP-RCC} & \multirow{3}{*}{(32,64,128,256)} & Upper & (150k,200k) & 210k & \multirow{2}{*}{1e-1} & \multirow{3}{*}{1e-4} \\
 &  &  & IB & (150k,200k) & 210k &  &  \\
 &  &  & UB & (50k,100k) & 150k & 1e-2 &  \\ \bottomrule
\end{tabular}

\label{tab:optimization}
\vspace{-5mm}
\end{table*}

\subsubsection{Optimization}
SGD optimizer is used with a momentum of 0.9, other paramters include the learning rate, milestones, total iteration and weight decay for each experiment can be seen in Table~\ref{tab:optimization}.
We first train the framework only with the identity-branch, then we load the pre-trained model and train the whole framework with the additional uncertainty-branch to generate the variance feature as well as optimize the mean feature simultaneously. 
The margin in each triplet loss is set to 0.2~\cite{hermans2017defense}. 
%

\subsubsection{Baselines}
To compare the effectiveness of our framework, we compare our method with several baselines as followings:


(1) \textit{RCC-Base}. It gives the results with the baseline model on CASIA-BN-RCC and OUMVLP-RCC with the original framework.

(2) \textit{RCC-Ours}. It optimizes the model on our benchmarks with our Progressive Feature Learning framework.

\subsection{Performance Comparison}
In this section, we show the performance of our framework on our constructed benchmarks.\\

\subsubsection{CASIA-BN-RCC}
The performance comparison on CASIA-BN-RCC is shown in Table~\ref{tab:casia-bn-sep}. 
%
%
The probe has been divided into three subsets according to the walking conditions and has been evaluated separately. 
It should be noted that since our framework aims at improving the precision in realistic cloth-changing condition, we take \textbf{the accuracy for CL as the main criteria}, since it is the hardest target in our task.
%
%
That is, it is necessary to improve the accuracy in CL condition when we could only extract cross-view and cross-cloth features from two non-overlapping datasets.
The characteristics of datasets will weaken the ability of baseline model to combine information.
However, our framework can bring large improvement for the baseline in both backbones for the CL conditions (+6.2\% for GaitSet, +3.2\% for GaitGL), indicating we can handle the more challenging cloth-changing problem. 
The accuracy of NM and BG also have some improvements compared with baseline, and the results with our framework in these two conditions are also reasonable.

\subsubsection{OUMVLP-RCC}
The performance comparison on OUMVLP-RCC is shown in Table~\ref{tab:oumvlp-sep}.
Notice that we also divide the probe into two subsets according to the walking conditions: NM and CL, and they have been evaluated separately, and we \textbf{take the accuracy for CL as the main criteria}. 
From the CL baseline results, we can learn that when adding a new cloth condition, the baseline model is difficult to learn cross-view and cross-cloth information from different identities, same with the result on CASIA-BN-RCC.
And our framework can further improve the model's performance in CL condition with both backbones (+2.9\% for GaitSet and +1.9\% for GaitGL), which indicate our framework can have reasonable performance on the hardest task.
And since the baseline's NM accuracy is near the upper bound, the improvement is not obvious, but it is comparable.

\subsection{Ablation Study}
To prove our framework is effective, we conducted several ablation studies with various settings on CASIA-BN-RCC and OUMVLP-RCC systematically. 
For simplicity, we only use the backbone of GaitSet as our backbone. \\

 

\subsubsection{Impact of Progressive Mapping}
We compare the performance of our Progressive Mapping with another three settings and show the results in Tabel~\ref{tab:sep}, to show the advance of our design.
In the first row, we show the performance of the one-stage module, which is also the baseline of our task.
All the other experiments employ the identity-branch and are equipped with the two-stage module, the difference lies in the triplet loss.
In the other three settings, we choose to use the normal triplet loss $\mathcal{L}_{n,j},\ j \in v,c$ behind CVM/CCM, and normal triplets are selected from $\mathcal{X}_v$ and $\mathcal{X}_c$.
\begin{equation}
\begin{aligned}
  \mathcal{L} _{nv,i}=&\frac{1}{|\mathcal{T}_v |}\sum_{(a,p,n)\in \mathcal{T}_v}{\left[ \left\| \mu _{v,i}^a-\mu _{v,i}^p \right\| ^2-\left\| \mu _{v,i}^a-\mu _{v,i}^n \right\| ^2+m \right] _+}  
    \\ 
  \mathcal{L} _{nc,i}=&\frac{1}{|\mathcal{T}_c |}\sum_{(a,p,n)\in \mathcal{T}_c}{\left[ \left\| \mu _{c,i}^a-\mu _{c,i}^p \right\| ^2-\left\| \mu _{c,i}^a-\mu _{c,i}^n \right\| ^2+m \right] _+} 
\end{aligned}
\end{equation}

In the experiment of the second line, we only add $\mathcal{L}_{nc,i}$ behind CCM, and the result shows that only constrain the two-stage module at the end of the final stage does not enough.
Compared with our methods, the results in the third row are also slightly poorer, for the negative repel effect of $\mathcal{L}_{nv,i}$ enlarges the intra-class distance of samples in $\mathcal{X}_c$, which is larger than the positive aggregate effect of $\mathcal{L}_{nc,i}$ since the diversity is richer in $\mathcal{X}_v$, so the adversarial effect will add difficulty to the optimization of CCM. 
With the four experiments, we can demonstrate that our improvement does not lie in the increased module parameter but in the two-stage module and the design of the triplets, making the CL condition has improvement (GaitSet: +3.3\%, GaitGL: + 1.2\%). 
It's important to recognize that our approach is not solely focused on achieving a small increase in accuracy; rather, it is can be scalable with additional data and computing resources, allowing for the potential to achieve even greater levels of accuracy.

In Table~\ref{tab:model}, we show that extending the baseline to a two-stage framework does not add too many parameters. The advantage from improved accuracy outweighs the larger architecture.

\begin{table}[t]
\begin{center}
		\caption{The model size with each backbone on CAISA-BNN-RCC.}
\begin{tabular}{ccc}
\hline
\multirow{2}{*}{Backbone} & \multicolumn{2}{c}{Model Size}             \\ \cline{2-3} 
                          & One-Stage & Two-Stage \\ \hline
GaitSet                   &  0.139M          &  0.205M                  \\
GaitGL                    &  0.428M          &  0.494M                  \\ \hline
\end{tabular}
\label{tab:model}
\end{center}

\end{table}
\subsubsection{Impact of Progressive Uncertainty}
We show the effect of Progressive Uncertainty in Table~\ref{tab:embedding}. 
The first row shows the result when training with the uncertainty-branch and using the Uncertainty-ware Triplet Loss $\mathcal{L}_{tv,i}$,\ $\mathcal{L}_{tc,i}$.
The second row shows the result that we re-sampled one embedding through $\mathcal{N}\left(\mu_{j,i}, \sigma_{j,i}^{2} \mathrm{I}\right)$ separately.
Our results demonstrate that only using the variance to help optimize the gradient can already improve the performance in the CL condition. 
That means the performance of our Progressive Uncertainty method mostly depends on the Uncertainty-aware Triplet Loss.
In the third and fourth rows, we only calculate $\sigma_{c,i}$, $\mathcal{L}_{tc,i}$, and still map the sequence as a point behind CVM. 
The difference is that in the third row, we only use CCM to generate the $\sigma_{c,i}$, and in the fourth row, we use both CVM and CCM to generate $\sigma_{c,i}$, but we abandon the residual connection between CVM and CCM. 
Our results demonstrate that only mapping each sequence as one Gaussian distribution cannot have the best performance.

\begin{table}[t]
\centering
		\caption{Progressive Uncertainty Performance in our framework. The results are reported in the rank-1 accuracy. \textit{e} represent that \textit{re-sample an additional embedding from Gaussian distributions}. Blue color indicates \textit{the comprehensive best result.} $^\ddag$ means we abandon the residual connection.}

\setlength{\tabcolsep}{0.8mm}
\resizebox{\columnwidth}{!}{%
		\begin{tabular}{cccccc}
			\hline
			Dataset                   & \multicolumn{3}{c}{CASIA-BN-RCC}  & \multicolumn{2}{c}{OUMVLP-RCC}  \\ \cline{2-6} 
			Settings        & NM $\uparrow$       & BG $\uparrow$    & CL $\uparrow$     & NM $\uparrow$  & CL $\uparrow$ \\ \hline
			$\sigma_{v,i}$ + $\sigma_{c,i}$  & 90.1  & 82.8  & \textbf{51.6}   &  85.0          & 57.3    \\
			$\sigma_{v,i}$ + $\sigma_{c,i}$ + \textit{e} & \textcolor{RoyalBlue}{90.2} & \textcolor{RoyalBlue}{83.1}  & \textcolor{RoyalBlue}{51.4}   & \textcolor{RoyalBlue}{85.0}           &  \textbf{57.3}  \\
			$\sigma_{c,i}$ (CCM) + \textit{e}   & \textbf{90.3}  & \textbf{83.2}   & 50.7   &   85.2             &  56.6  \\
			{\begin{tabular}[c]{@{}c@{}}$\sigma_{c,i}$ (CVM + CCM)$^\ddag$ + \textit{e} \end{tabular}}  & 90.2   & 83.0   & 50.3   &   \textbf{85.2}             & 56.6   \\ \hline
		\end{tabular}
		\label{tab:embedding}    
}
\end{table}



%
We further provide the analysis of progressive feature learning, illustrate the limitations of our method, and introduce future work in the supplementary materials.

\section{Conclusion}
\label{sec_conclusion}
In this work, we tackle the cloth-changing gait recognition problem in realistic application with automatic data collection at one of the first time. 
We propose a new framework that can be applied with off-the-shelf gait recognition backbones to boost performance in the RCC-GR task.
%
%
The key methods of our Progressive Feature Learning framework are that we design Progressive Mapping and Progressive Uncertantity to span the cross-view and cross-cloth feature in a cascaded way.
Extensive experiments were conducted to validate the effectiveness of our framework. 
It shows that our framework could improve the performance of state-of-the-art methods in walking with different coats (CL) condition in RCC-GR.

\bibliographystyle{IEEEtran}
\bibliography{IEEEabrv,egbib.bib}

\section{Biography Section}

\begin{IEEEbiographynophoto}{Xuqian Ren}
received the B.E. degree from University of Science and Technology Beijing in 2019, received the M.S. degree from Beijing Institute of Technology in 2022. She is currently a Ph.D. candidate with Computer Science Unit, Faculty of Information Technology and Communication Sciences, Tampere Universities, Finland. Her current research interests include contrastive learning, image generation and 3d reconstruction.
\end{IEEEbiographynophoto}

\vspace{-5mm}
\begin{IEEEbiographynophoto}{Saihui Hou}
received the B.E. and Ph.D. degrees from University of Science and Technology of China in 2014 and 2019, respectively. He is currently an Assistant Professor with School of Artificial Intelligence, Beijing Normal University, and works in cooperation with Watrix Technology Limited Co. Ltd. His research interests include computer vision and machine learning. He recently focuses on gait recognition which aims to identify different people according to the walking patterns.
\end{IEEEbiographynophoto}

\vspace{-5mm}
\begin{IEEEbiographynophoto}{Chunshui Cao}
received the B.E. and Ph.D. degrees from University of Science and Technology of China in 2013 and 2018, respectively. During his Ph.D. study, he joined Center for Research on Intelligent Perception and Computing, National Laboratory of Pattern Recognition, Institute of Automation, Chinese Academy of Sciences. From 2018 to 2020, he worked as a Postdoctoral Fellow with PBC School of Finance, Tsinghua University. He is currently a Research Scientist with Watrix Technology Limited Co. Ltd. His research interests include pattern recognition, computer vision and machine learning.
\end{IEEEbiographynophoto}
\vspace{-5mm}
\begin{IEEEbiographynophoto}{Xu Liu}
received the B.E. and Ph.D. degrees from University of Science and Technology of China in 2013 and 2018, respectively. He is currently a Research Scientist with Watrix Technology Limited Co. Ltd. His research interests include gait recognition, object detection and image segmentation.
\end{IEEEbiographynophoto}
\vspace{-5mm}
\begin{IEEEbiographynophoto}{Yongzhen Huang}
received the B.E. degree from Huazhong University of Science and Technology in 2006, and the Ph.D. degree from Institute of Automation, Chinese Academy of Sciences in 2011. He is currently an Associate Professor with School of Artificial Intelligence, Beijing Normal University, and works in cooperation with Watrix Technology Limited Co. Ltd. He has published one book and more than 80 papers at international journals and conferences such as TPAMI, IJCV, TIP, TSMCB, TMM, TCSVT, CVPR, ICCV, ECCV, NIPS, AAAI. His research interests include pattern recognition, computer vision and machine learning.
\end{IEEEbiographynophoto}

\vfill

\end{document}


\title{Supplementary Materials of \\ Progressive Feature Learning for \\ Realistic Cloth-Changing Gait Recognition}

\markboth{Journal of \LaTeX\ Class Files,~Vol.~14, No.~8, August~2021}%
{Shell \MakeLowercase{\textit{et al.}}: A Sample Article Using IEEEtran.cls for IEEE Journals}

\IEEEpubid{0000--0000/00\$00.00~\copyright~2021 IEEE}

\maketitle

\section{Analysis about Progressive Feature learning}
In this section, we discuss why Progressive Mapping and Progressive Uncertainty can boost the performance of the walking with different coats (CL) condition. 

The first explanation is that our two methods can optimize the learning direction of our network.
Based on the data ratio in our task, the benchmarks are long-tailed.
So, in our methods, our Progressive-aware Triplets can make the two-stage module first learn the features from $\mathcal{X}_v$ (head class), then learn to optimize the feature from $\mathcal{X}_c$ (tail class) as well as maintain the ability to classify head class features. 
And in this way, the feature can mainly be extracted from the head classes, learning a more robust model for the identity classification task. 

\begin{figure}[t]
	\centering	
	\includegraphics[width=\columnwidth]{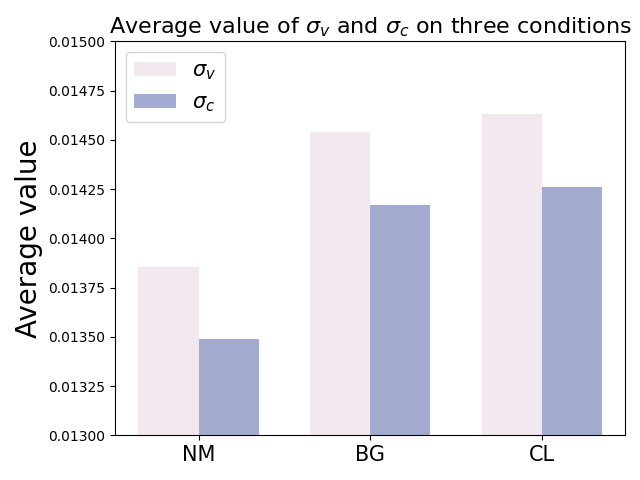}		
        \caption{The average $\sigma_v$ and $\sigma_c$ of each condition.}
         \label{fig:sigma}
	
 \end{figure}

Furthermore, Progressive Uncertainty can improve the weight of the head class features in the gradient during training. 
The head classes have richer intra-class diversity, so they are more easily optimized and can contribute more to the gradient at the beginning of the training process, while the tail class lacks intra-class diversity, so they are hard samples and are not easy to learn and optimize. 
This property will make the model generate smaller variance for classes and parts which can be easily learned and will generate larger variance for classes and parts which are harder to learn. 
The variance we generated for each sequence is put in the denominator position of the triplet loss, so the triplet loss is uncertainty aware. 
The feature with smaller variance will have a larger gradient during training, making the network focus more on these features. 
While if these head class features are well optimized, the network will not be misled by the tail class and will learn a more robust model. 
Figure~\ref{fig:sigma} shows the average variance in the test dataset for the three conditions in CASIA-BN-RCC separately.

\begin{figure}[t]
        \centering	
	\includegraphics[width=\columnwidth]{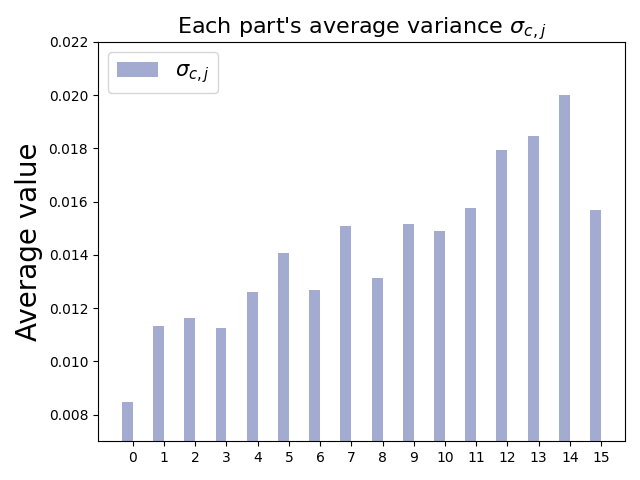}	
	\caption{The variance of the different parts of features averaged along the feature dimension axis.}
	\label{fig:parts}

\end{figure}

We denote the average of 16 different parts as $\sigma_{v/c}$, where each part is represented by $\sigma_{v/c,j}$.
We can see that the variance of the easier optimization class NM is relatively smaller than the BG, CL condition.
Also, the $\sigma_c$ after CCM is smaller than the $\sigma_v$ after CVM. 
Figure~\ref{fig:parts} illustrates $\sigma_{c,j}$ averaged along the feature dimension axis, which represents the standard deviation of the 16 different parts of features.

\begin{figure*}[ht]    
        \centering	
        
	\includegraphics[width=\textwidth]{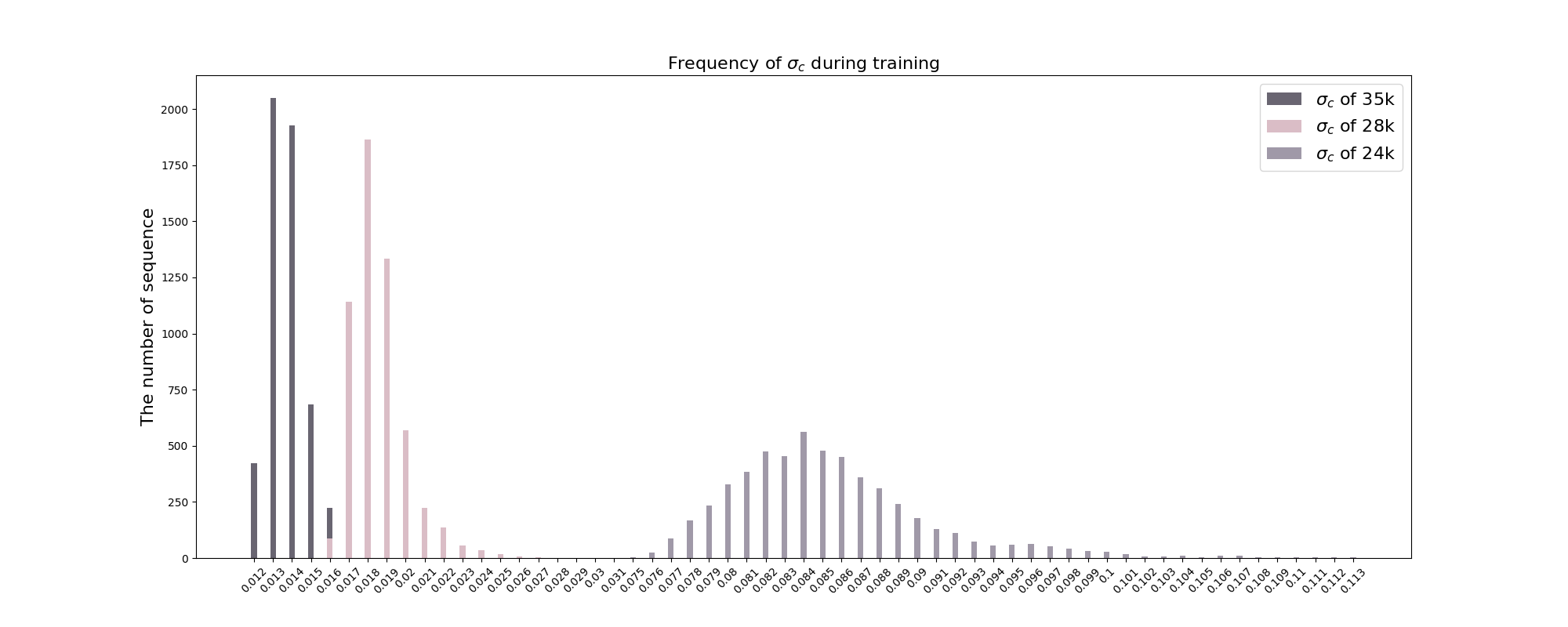}	
        \caption{Changes of the average $\sigma_c$ during training. During training, the number of feature that has large variance become less, which means better feature has been extracted.}	
        \label{fig:process}
        
 \end{figure*}

The two pictures indicate that the variance can represent the uncertainty of the classification result of each stage. 
For classes with more varieties, such as CL condition, they have larger variance, which means they are hard to learn and have large uncertainty and low confidence. 
For parts in silhouettes, such as the legs parts (11-15 parts in the feature), their dynamic changes are more obvious, so they have larger uncertainty and larger variance. 
Therefore, our variance can model the uncertainty in both distributions (class type) level and feature (part) level. 
Also, in Figure~\ref{fig:process}, we can see that during training, the $\sigma_c$ decreases gradually. 
In conclusion, the smaller the variance, the better the sequence feature has been learned.

\IEEEpubidadjcol
The second explanation is that by applying SVD decomposition for the CCM and reconstructing the CCM only use the first eigenvalue and its corresponding eigenvector, we find the essence of the cross-cloth problem is that the feature space has been compressed towards the direction of the centerline of the NM to CL feature projection. 

\begin{figure}[t]
        \centering	
	\includegraphics[width=\columnwidth]{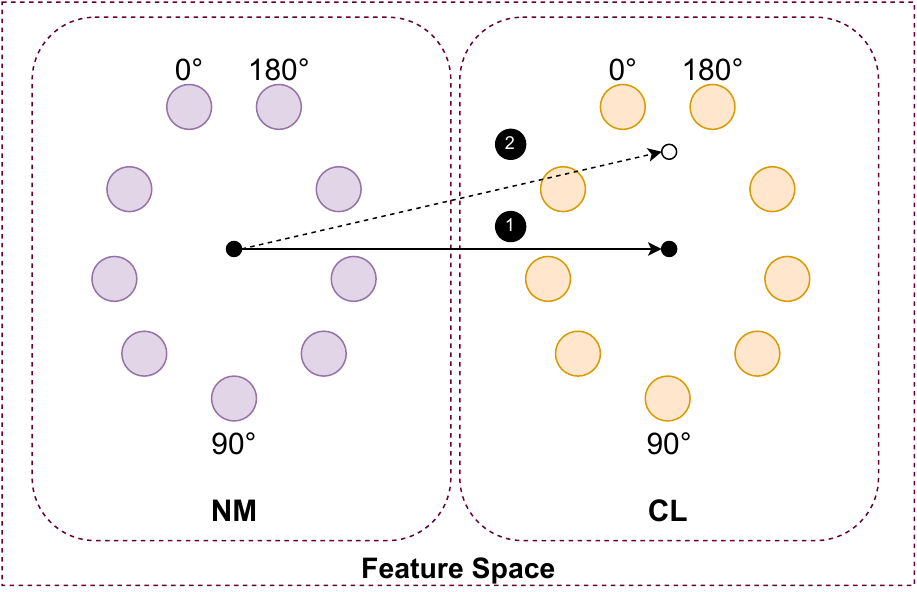}	
	\caption{The feature distribution of NM and CL in feature space. When the sequences in CL condition are rich, the composition direction is along line 1. However, when CL only has sequences in front views, the center point of CL features changes, and the composition direction moves to line 2, which is inaccurate.}
	\label{fig:yasuo}
 \end{figure}

When the dataset is rich, we can learn a more accurate direction to compress the feature space by CCM, but when the dataset lack information, especially when the CL condition only has views limited in several perspectives, the center point of the CL feature of each identity will not be accurate, as shown in Figure~\ref{fig:yasuo}.
 
Since our algorithm can learn a more robust compress direction towards NM to CL, the performance on CL condition can be improved.

\section{Limitations and future work}
In this work, we make one of the first attempts toward the Realistic Cloth-Changing Gait Recognition task that train gait recognition model on automatic labeled datasets, which is more practical and meets the industry requirements. 
Although we have explored the effectiveness of our framework with Progressive Feature Learning, there are still some limitations in our work that need further exploration.
First, our work focuses on the realistic cloth-changing problem. 
To support our research, we construct two benchmarks from existing datasets to simulate the real scene. 
In fact, with enough funds and time, better and larger datasets collected from the real world can be constructed to further help our task.
Second, our work aims to solve the model over-fitting and bias caused by the long-tail property of datasets. However, based on our framework, there is still a large development space for the CL condition to improve. 
So further research can focus on developing better frameworks to improve the accuracy of CL condition continuously.

The necessity of our work is to propose the RCC-GR task, which is a very practical and common problem when applying gait recognition in the industry. At present, the application of gait recognition in academics is limited to using manually annotated datasets, which can be time-consuming and expensive. The proposal of the RCC-GR task enables the training of models on automatically labeled datasets, making gait recognition more practical and economical.

\vfill

%% file: CASIA-BN2-experiment.tex
\begin{table*}[t]
\begin{center}
	\caption{The rank-1 accuracy (\%) on CASIA-BN-RCC for different probe views excluding the identical-view cases. For evaluation, the sequences of NM-01,02,03,04 for each subject are taken as the gallery. The probe sequences are divided into three subsets according to the walking conditions (\textit{i.e.} NM, BG, CL).}
            \setlength{\tabcolsep}{1.4mm}
		\begin{tabular}{ccccccccccccccc}
			\toprule
			\multirow{2}{*}{Backbone} &
			\multirow{2}{*}{Condition} &
			\multicolumn{1}{c}{\multirow{2}{*}{Method}} &
			\multicolumn{11}{c}{Probe View} &
			\multirow{2}{*}{Average} \\ \cline{4-14}
			&
			&
			\multicolumn{1}{c}{} & \multicolumn{1}{c}{$0^{\circ}$} & \multicolumn{1}{c}{$18^{\circ}$} & \multicolumn{1}{c}{$36^{\circ}$} &
			\multicolumn{1}{c}{$54^{\circ}$} & \multicolumn{1}{c}{$72^{\circ}$} & \multicolumn{1}{c}{$90^{\circ}$} &
			\multicolumn{1}{c}{$108^{\circ}$} & \multicolumn{1}{c}{$126^{\circ}$} & \multicolumn{1}{c}{$134^{\circ}$} &
			\multicolumn{1}{c}{$162^{\circ}$} & \multicolumn{1}{c}{$180^{\circ}$} & \\ \midrule
					
			\multirow{6}{*}{GaitSet} &
			\multirow{2}{*}{NM-5,6} 
			&
			\multicolumn{1}{c}{RCC-Base} 
			&84.1 & 91.3 & 98.4 &
			94.5 & 86.3 & 83.1 &
			86.9 & 93.0 & 97.9 &
			92.4 & 78.5 & 89.7 \\			
			&
			&
			\multicolumn{1}{c}{RCC-Ours} &
			84.3 & 93.1 & 97.3 &
			94.3 & 87.6 & 82.7 &
			87.6 & 93.4 & 97.2 &
			93.3 & 81.7 & \textbf{90.2} \\ 
			  \cline{2-15}
			&
			\multirow{2}{*}{BG-1,2} 
			&
			\multicolumn{1}{c}{RCC-Base} &
			79.6 & 86.6 & 91.7 &
			86.2 & 77.2 & 71.2 &
			76.7 & 81.5 & 89.4 &
			87.7 & 74.0 & 82.0 \\
			&
			& 
			\multicolumn{1}{c}{RCC-Ours} &
			78.5 & 86.0 & 91.8 &
			87.5 & 78.9 & 75.5 &
			79.8 & 83.2 & 90.6 &
			87.6 & 74.6 & \textbf{83.1} \\ \cline{2-15}
			&
			\multirow{2}{*}{CL-1,2} 
			&
			\multicolumn{1}{c}{RCC-Base} &
			43.5 & 53.2 & 54.8 &
			47.4 & 45.1 & 36.3 &
			36.1 & 43.3 & 50.6 &
			49.3 & 37.5 & 45.2           \\  
			&
			&
			\multicolumn{1}{c}{RCC-Ours} & 
			\textbf{50.7} & \textbf{60.9} & \textbf{61.9} & 
			\textbf{55.0} & \textbf{47.9} & \textbf{40.3} &
			\textbf{45.5} & \textbf{53.1} & \textbf{55.2} &  
			\textbf{56.4} & \textbf{38.4} & \textbf{51.4} \\ \midrule

			\multirow{6}{*}{GaitGL} &
			\multirow{2}{*}{NM-5,6} 
			&
			\multicolumn{1}{c}{RCC-Base} &
			88.0 & 94.1 & 96.9 &
			94.3 & 86.8 & 82.8 &
			89.2 & 94.5 & 97.4 &
			93.3 & 79.4 & 90.6 \\
			&
			&
			\multicolumn{1}{c}{RCC-Ours} &
			87.4 & 93.5 & 96.6 &
			94.4 & 87.7 & 82.6 &
			89.6 & 94.8 & 97.3 &
			92.0 & 80.3 & \textbf{90.6} \\ \cline{2-15} 
			&
			\multirow{2}{*}{BG-1,2} 
			&
			\multicolumn{1}{c}{RCC-Base} &
			81.0 & 89.1 & 94.2 &
			88.7 & 79.6 & 73.1 &
			78.6 & 86.8 & 93.4 &
			89.4 & 68.5 & 83.9 \\
			&
			&
			\multicolumn{1}{c}{RCC-Ours} &
			81.7 & 89.3 & 93.7 &
			89.3 & 80.9 & 76.4 &
			79.2 & 88.5 & 92.9 &
			90.8 & 70.5 & \textbf{84.8} \\ \cline{2-15} 
			&
			\multirow{2}{*}{CL-1,2} 
			&
			\multicolumn{1}{c}{RCC-Base} &
			57.5 & 66.3 & 64.6 &
			59.5 & 59.5 & 52.7 &
			53.3 & 57.2 & 63.6 &
			59.8 & 39.7 & 57.6 \\
			&
			&
			\multicolumn{1}{c}{RCC-Ours} &
			\textbf{58.2} & \textbf{71.3} & \textbf{73.5} &
			\textbf{64.6} & \textbf{60.9} & \textbf{53.7} &
			\textbf{54.5} & \textbf{60.4} & \textbf{68.1} &
			\textbf{65.6} & 38.0 & \textbf{60.8} \\ \bottomrule
		\end{tabular}%
	\label{tab:casia-bn-sep}
\end{center}
\end{table*}

%% file: OUMVLP-experiment.tex
\begin{table*}[t]
\begin{center}
	\caption{The rank-1 accuracy (\%) on OUMVLP-RCC for different probe views excluding the identical-view cases. For evaluation, the sequences of NM-01 for each subject are taken as the gallery and the probe sequences of the two walking conditions, NM-00 and CL-00, are respectively evaluated.}
        \setlength{\tabcolsep}{0.6mm}
		\begin{tabular}{cccccccccccccccccc}
			\toprule
			\multirow{2}{*}{Backbone} &
			\multirow{2}{*}{Condition} &
			\multirow{2}{*}{Method} &
			\multicolumn{14}{c}{Probe View} &
			\multirow{2}{*}{Average} \\ \cline{4-17}
			
                &
			&
			&
			$0^{\circ}$ & $15^{\circ}$ &$30^{\circ}$ &
			$45^{\circ}$ & $60^{\circ}$ &$75^{\circ}$ &
			$90^{\circ}$ & $180^{\circ}$ &$195^{\circ}$ &
			$210^{\circ}$ &$225^{\circ}$ &$240^{\circ}$ &
			$255^{\circ}$ & $270^{\circ}$ &
			\\ \midrule
			\multirow{4}{*}{GaitSet} &
			\multirow{2}{*}{NM-00} 
			&
			RCC-Base &
			76.4 & 85.4 & 88.7 &
			88.8 & 84.2 & 86.3 &
			84.4 & 78.9 & 84.1 &
			87.8 & 87.9 & 83.8 &
			85.2 & 82.8 & 84.6\\
			&
			&
			RCC-Ours &
			76.8 & 85.9 & 89.1 &
			89.3 &
			84.8 & 86.7 & 84.8 &
			79.1 & 84.4 & 88.0 &
			88.2 & 84.3 & 85.8 &
			83.2 &  \textbf{85.0}\\ \cline{2-18} 
			&
			\multirow{2}{*}{CL-00} 
			&
			RCC-Base &
			66.9 & 78.1 & 79.6 &
			65.0 & 30.1 & 31.1 &
			22.3 & 70.2 & 77.4 &
			80.4 & 70.2 & 36.1 &
			32.5 & 21.8 & 54.4\\
			&
			&
			RCC-Ours &
			\textbf{67.2} & \textbf{78.3} & \textbf{80.9} &
			\textbf{68.6} & \textbf{35.5} & \textbf{35.7} &
			\textbf{26.4} & \textbf{70.3} & \textbf{77.7} &
			\textbf{81.5} & \textbf{73.2} & \textbf{41.8} &
			\textbf{38.3} & \textbf{27.0} & \textbf{57.3} \\ \midrule
			
			\multirow{4}{*}{GaitGL} &
			\multirow{2}{*}{NM-00} 
			&
			RCC-Base &
			86.5 & 90.4 & 91.1 &
			91.3 & 90.6 & 90.5 &
			90.0 & 89.5 & 89.0 &
			90.2 & 90.2 & 89.3 &
			89.2 & 88.6 & 89.7\\
			&
			&
			RCC-Ours &
			86.6 & 90.4 & 91.9 &
			91.4 & 90.8 & 90.7 &
			90.2 & 89.7 & 88.8 &
			90.2 & 90.4 & 89.5 &
			89.4 & 88.8 & \textbf{89.8}\\ \cline{2-18} 
			&
			\multirow{2}{*}{CL-00} 
			&
			RCC-Base &
			83.1 & 88.4 & 88.9 &
			85.5 & 67.7 & 67.5 &
			58.6 & 86.9 & 86.9 &
			88.6 & 86.1 & 69.5 &
			67.0 & 55.8 & 77.2\\
			&
			&
			RCC-Ours &
			\textbf{83.3} & \textbf{88.2} & \textbf{89.1} &
			\textbf{86.6} & \textbf{71.6} & \textbf{72.8} &
			\textbf{65.8} & \textbf{87.0} & 86.4 &
			\textbf{88.7} & \textbf{86.8} & \textbf{73.0} &
			\textbf{69.1} & \textbf{59.5} & \textbf{79.1}\\ \bottomrule
			
		\end{tabular}%
   
	\label{tab:oumvlp-sep}
\end{center}
\end{table*}

%% file: bare_jrnl_new_sample4.bbl
\begin{thebibliography}{10}
\providecommand{\url}[1]{#1}
\csname url@samestyle\endcsname
\providecommand{\newblock}{\relax}
\providecommand{\bibinfo}[2]{#2}
\providecommand{\BIBentrySTDinterwordspacing}{\spaceskip=0pt\relax}
\providecommand{\BIBentryALTinterwordstretchfactor}{4}
\providecommand{\BIBentryALTinterwordspacing}{\spaceskip=\fontdimen2\font plus
\BIBentryALTinterwordstretchfactor\fontdimen3\font minus
  \fontdimen4\font\relax}
\providecommand{\BIBforeignlanguage}[2]{{%
\expandafter\ifx\csname l@#1\endcsname\relax
\typeout{** WARNING: IEEEtran.bst: No hyphenation pattern has been}%
\typeout{** loaded for the language `#1'. Using the pattern for}%
\typeout{** the default language instead.}%
\else
\language=\csname l@#1\endcsname
\fi
#2}}
\providecommand{\BIBdecl}{\relax}
\BIBdecl

\bibitem{xu2021long}
P.~Xu and X.~Zhu, ``Deepchange: A large long-term person re-identification
  benchmark with clothes change,'' \emph{arXiv preprint arXiv:2105.14685},
  2021.

\bibitem{larsen2008gait}
P.~K. Larsen, E.~B. Simonsen, and N.~Lynnerup, ``Gait analysis in forensic
  medicine,'' \emph{Journal of forensic sciences}, vol.~53, no.~5, pp.
  1149--1153, 2008.

\bibitem{bouchrika2011using}
I.~Bouchrika, M.~Goffredo, J.~Carter, and M.~Nixon, ``On using gait in forensic
  biometrics,'' \emph{Journal of forensic sciences}, vol.~56, no.~4, pp.
  882--889, 2011.

\bibitem{black2017forensic}
S.~Black, M.~Wall, R.~Abboud, R.~Baker, and J.~Stebbins, ``Forensic gait
  analysis: A primer for courts,'' 2017.

\bibitem{yu2006framework}
S.~Yu, D.~Tan, and T.~Tan, ``A framework for evaluating the effect of view
  angle, clothing and carrying condition on gait recognition,'' in \emph{18th
  international conference on pattern recognition (ICPR'06)}, vol.~4.\hskip 1em
  plus 0.5em minus 0.4em\relax IEEE, 2006, pp. 441--444.

\bibitem{liao2017pose}
R.~Liao, C.~Cao, E.~B. Garcia, S.~Yu, and Y.~Huang, ``Pose-based
  temporal-spatial network (ptsn) for gait recognition with carrying and
  clothing variations,'' in \emph{Chinese Conference on biometric
  recognition}.\hskip 1em plus 0.5em minus 0.4em\relax Springer, 2017, pp.
  474--483.

\bibitem{takemura2018multi}
N.~Takemura, Y.~Makihara, D.~Muramatsu, T.~Echigo, and Y.~Yagi, ``Multi-view
  large population gait dataset and its performance evaluation for cross-view
  gait recognition,'' \emph{IPSJ Transactions on Computer Vision and
  Applications}, vol.~10, no.~1, pp. 1--14, 2018.

\bibitem{connor2018biometric}
P.~Connor and A.~Ross, ``Biometric recognition by gait: A survey of modalities
  and features,'' \emph{Computer Vision and Image Understanding}, vol. 167, pp.
  1--27, 2018.

\bibitem{9359526}
A.~Zhao, J.~Dong, J.~Li, L.~Qi, and H.~Zhou, ``Associated spatio-temporal
  capsule network for gait recognition,'' \emph{IEEE Transactions on
  Multimedia}, vol.~24, pp. 846--860, 2022.

\bibitem{zhu2021gait}
Z.~Zhu, X.~Guo, T.~Yang, J.~Huang, J.~Deng, G.~Huang, D.~Du, J.~Lu, and
  J.~Zhou, ``Gait recognition in the wild: A benchmark,'' in \emph{Proceedings
  of the IEEE/CVF International Conference on Computer Vision}, 2021, pp.
  14\,789--14\,799.

\bibitem{chen2017person}
Y.-C. Chen, X.~Zhu, W.-S. Zheng, and J.-H. Lai, ``Person re-identification by
  camera correlation aware feature augmentation,'' \emph{IEEE transactions on
  pattern analysis and machine intelligence}, vol.~40, no.~2, pp. 392--408,
  2017.

\bibitem{zheng2016person}
L.~Zheng, Y.~Yang, and A.~G. Hauptmann, ``Person re-identification: Past,
  present and future,'' \emph{arXiv preprint arXiv:1610.02984}, 2016.

\bibitem{deng2019arcface}
J.~Deng, J.~Guo, N.~Xue, and S.~Zafeiriou, ``Arcface: Additive angular margin
  loss for deep face recognition,'' in \emph{Proceedings of the IEEE/CVF
  conference on computer vision and pattern recognition}, 2019, pp. 4690--4699.

\bibitem{deng2020retinaface}
J.~Deng, J.~Guo, E.~Ververas, I.~Kotsia, and S.~Zafeiriou, ``Retinaface:
  Single-shot multi-level face localisation in the wild,'' in \emph{Proceedings
  of the IEEE/CVF conference on computer vision and pattern recognition}, 2020,
  pp. 5203--5212.

\bibitem{liu2020deep}
J.~Liu, Y.~Sun, C.~Han, Z.~Dou, and W.~Li, ``Deep representation learning on
  long-tailed data: A learnable embedding augmentation perspective,'' in
  \emph{Proceedings of the IEEE/CVF conference on computer vision and pattern
  recognition}, 2020, pp. 2970--2979.

\bibitem{liao2020model}
R.~Liao, S.~Yu, W.~An, and Y.~Huang, ``A model-based gait recognition method
  with body pose and human prior knowledge,'' \emph{Pattern Recognition},
  vol.~98, p. 107069, 2020.

\bibitem{li2020end}
X.~Li, Y.~Makihara, C.~Xu, Y.~Yagi, S.~Yu, and M.~Ren, ``End-to-end model-based
  gait recognition,'' in \emph{Proceedings of the Asian conference on computer
  vision}, 2020.

\bibitem{li2021end}
X.~Li, Y.~Makihara, C.~Xu, and Y.~Yagi, ``End-to-end model-based gait
  recognition using synchronized multi-view pose constraint,'' in
  \emph{Proceedings of the IEEE/CVF International Conference on Computer
  Vision}, 2021, pp. 4106--4115.

\bibitem{li2022strong}
N.~Li and X.~Zhao, ``A strong and robust skeleton-based gait recognition method
  with gait periodicity priors,'' \emph{IEEE Transactions on Multimedia}, pp.
  1--1, 2022.

\bibitem{9478261}
K.~Xu, X.~Jiang, and T.~Sun, ``Gait recognition based on local graphical
  skeleton descriptor with pairwise similarity network,'' \emph{IEEE
  Transactions on Multimedia}, vol.~24, pp. 3265--3275, 2022.

\bibitem{shiraga2016geinet}
K.~Shiraga, Y.~Makihara, D.~Muramatsu, T.~Echigo, and Y.~Yagi, ``Geinet:
  View-invariant gait recognition using a convolutional neural network,'' in
  \emph{2016 international conference on biometrics (ICB)}.\hskip 1em plus
  0.5em minus 0.4em\relax IEEE, 2016, pp. 1--8.

\bibitem{hu2018robust}
B.~Hu, Y.~Gao, Y.~Guan, Y.~Long, N.~Lane, and T.~Ploetz, ``Robust cross-view
  gait identification with evidence: A discriminant gait gan (diggan) approach
  on 10000 people,'' \emph{arXiv e-prints}, pp. arXiv--1811, 2018.

\bibitem{he2018multi}
Y.~He, J.~Zhang, H.~Shan, and L.~Wang, ``Multi-task gans for view-specific
  feature learning in gait recognition,'' \emph{IEEE Transactions on
  Information Forensics and Security}, vol.~14, no.~1, pp. 102--113, 2018.

\bibitem{liu2018learning}
W.~Liu, C.~Zhang, H.~Ma, and S.~Li, ``Learning efficient spatial-temporal gait
  features with deep learning for human identification,''
  \emph{Neuroinformatics}, vol.~16, no.~3, pp. 457--471, 2018.

\bibitem{wolf2016multi}
T.~Wolf, M.~Babaee, and G.~Rigoll, ``Multi-view gait recognition using 3d
  convolutional neural networks,'' in \emph{2016 IEEE international conference
  on image processing (ICIP)}.\hskip 1em plus 0.5em minus 0.4em\relax IEEE,
  2016, pp. 4165--4169.

\bibitem{8643814}
S.~Li, W.~Liu, and H.~Ma, ``Attentive spatial–temporal summary networks for
  feature learning in irregular gait recognition,'' \emph{IEEE Transactions on
  Multimedia}, vol.~21, no.~9, pp. 2361--2375, 2019.

\bibitem{lin2020gait}
B.~Lin, S.~Zhang, and F.~Bao, ``Gait recognition with multiple-temporal-scale
  3d convolutional neural network,'' in \emph{ACM MM}, 2020, pp. 3054--3062.

\bibitem{lin2021gait}
B.~Lin, S.~Zhang, and X.~Yu, ``Gait recognition via effective global-local
  feature representation and local temporal aggregation,'' in \emph{Proceedings
  of the IEEE/CVF International Conference on Computer Vision}, 2021, pp.
  14\,648--14\,656.

\bibitem{9767609}
L.~Yao, W.~Kusakunniran, P.~Zhang, Q.~Wu, and J.~Zhang, ``Improving
  disentangled representation learning for gait recognition using group
  supervision,'' \emph{IEEE Transactions on Multimedia}, pp. 1--1, 2022.

\bibitem{chao2019gaitset}
H.~Chao, Y.~He, J.~Zhang, and J.~Feng, ``Gaitset: Regarding gait as a set for
  cross-view gait recognition,'' in \emph{Proceedings of the AAAI conference on
  artificial intelligence}, vol.~33, no.~01, 2019, pp. 8126--8133.

\bibitem{fan2020gaitpart}
C.~Fan, Y.~Peng, C.~Cao, X.~Liu, S.~Hou, J.~Chi, Y.~Huang, Q.~Li, and Z.~He,
  ``Gaitpart: Temporal part-based model for gait recognition,'' in
  \emph{Proceedings of the IEEE/CVF conference on computer vision and pattern
  recognition}, 2020, pp. 14\,225--14\,233.

\bibitem{hou2020gait}
S.~Hou, C.~Cao, X.~Liu, and Y.~Huang, ``Gait lateral network: Learning
  discriminative and compact representations for gait recognition,'' in
  \emph{Computer Vision--ECCV 2020: 16th European Conference, Glasgow, UK,
  August 23--28, 2020, Proceedings, Part IX}.\hskip 1em plus 0.5em minus
  0.4em\relax Springer, 2020, pp. 382--398.

\bibitem{hou2021set}
S.~Hou, X.~Liu, C.~Cao, and Y.~Huang, ``Set residual network for
  silhouette-based gait recognition,'' \emph{IEEE Transactions on Biometrics,
  Behavior, and Identity Science}, 2021.

\bibitem{shi2019probabilistic}
Y.~Shi and A.~K. Jain, ``Probabilistic face embeddings,'' in \emph{Proceedings
  of the IEEE/CVF International Conference on Computer Vision}, 2019, pp.
  6902--6911.

\bibitem{yu2019robust}
T.~Yu, D.~Li, Y.~Yang, T.~M. Hospedales, and T.~Xiang, ``Robust person
  re-identification by modelling feature uncertainty,'' in \emph{Proceedings of
  the IEEE/CVF International Conference on Computer Vision}, 2019, pp.
  552--561.

\bibitem{9707634}
P.~Wang, C.~Ding, W.~Tan, M.~Gong, K.~Jia, and D.~Tao, ``Uncertainty-aware
  clustering for unsupervised domain adaptive object re-identification,''
  \emph{IEEE Transactions on Multimedia}, pp. 1--1, 2022.

\bibitem{9483643}
X.~Song and Z.~Jin, ``Robust label rectifying with consistent
  contrastive-learning for domain adaptive person re-identification,''
  \emph{IEEE Transactions on Multimedia}, vol.~24, pp. 3229--3239, 2022.

\bibitem{8693882}
X.~Yang, Y.~Gao, H.~Luo, C.~Liao, and K.-T. Cheng, ``Bayesian denet: Monocular
  depth prediction and frame-wise fusion with synchronized uncertainty,''
  \emph{IEEE Transactions on Multimedia}, vol.~21, no.~11, pp. 2701--2713,
  2019.

\bibitem{9439889}
D.~Guan, J.~Huang, A.~Xiao, S.~Lu, and Y.~Cao, ``Uncertainty-aware unsupervised
  domain adaptation in object detection,'' \emph{IEEE Transactions on
  Multimedia}, vol.~24, pp. 2502--2514, 2022.

\bibitem{9882310}
J.~Zhuo, S.~Wang, and Q.~Huang, ``Uncertainty modeling for robust domain
  adaptation under noisy environments,'' \emph{IEEE Transactions on
  Multimedia}, pp. 1--14, 2022.

\bibitem{9899753}
Y.~Cui, W.~Deng, X.~Xu, Z.~Liu, Z.~Liu, M.~Pietikäinen, and L.~Liu,
  ``Uncertainty-guided semi-supervised few-shot class-incremental learning with
  knowledge distillation,'' \emph{IEEE Transactions on Multimedia}, pp. 1--14,
  2022.

\bibitem{9966825}
J.~Hong, W.~Zhang, Z.~Feng, and W.~Zhang, ``Dual cross-attention for video
  object segmentation via uncertainty refinement,'' \emph{IEEE Transactions on
  Multimedia}, pp. 1--16, 2022.

\bibitem{chang2020data}
J.~Chang, Z.~Lan, C.~Cheng, and Y.~Wei, ``Data uncertainty learning in face
  recognition,'' in \emph{Proceedings of the IEEE/CVF conference on computer
  vision and pattern recognition}, 2020, pp. 5710--5719.

\bibitem{shi2020towards}
Y.~Shi, X.~Yu, K.~Sohn, M.~Chandraker, and A.~K. Jain, ``Towards universal
  representation learning for deep face recognition,'' in \emph{Proceedings of
  the IEEE/CVF conference on computer vision and pattern recognition}, 2020,
  pp. 6817--6826.

\bibitem{chen2021reliable}
K.~Chen, Q.~Lv, T.~Yi, and Z.~Yi, ``Reliable probabilistic face embeddings in
  the wild,'' \emph{arXiv preprint arXiv:2102.04075}, 2021.

\bibitem{shi2021spatial}
Y.~Shi, W.~Tian, H.~Ling, Z.~Li, and P.~Li, ``Spatial-wise and channel-wise
  feature uncertainty for occluded person re-identification,''
  \emph{Neurocomputing}, vol. 486, pp. 237--249, 2022.

\bibitem{kingma2013auto}
D.~P. Kingma and M.~Welling, ``Auto-encoding variational bayes,'' \emph{arXiv
  preprint arXiv:1312.6114}, 2013.

\bibitem{paszke2019pytorch}
A.~Paszke, S.~Gross, F.~Massa, A.~Lerer, J.~Bradbury, G.~Chanan, T.~Killeen,
  Z.~Lin, N.~Gimelshein, L.~Antiga \emph{et~al.}, ``Pytorch: An imperative
  style, high-performance deep learning library,'' \emph{Advances in neural
  information processing systems}, vol.~32, 2019.

\bibitem{hermans2017defense}
A.~Hermans, L.~Beyer, and B.~Leibe, ``In defense of the triplet loss for person
  re-identification,'' \emph{arXiv preprint arXiv:1703.07737}, 2017.

\end{thebibliography}
